\pdfoutput=1

\documentclass[11pt]{article}

\usepackage[final]{acl}

\usepackage{times}
\usepackage{latexsym}

\usepackage[T1]{fontenc}

\usepackage[utf8]{inputenc}

\usepackage{microtype}

\usepackage{inconsolata}

\usepackage{graphicx}

\usepackage{booktabs}
\usepackage{multirow}
\usepackage{caption}
\usepackage{adjustbox} 
\usepackage{array}
\usepackage{xspace}
\usepackage{cleveref}
\usepackage{tcolorbox}

\newcommand{\methodlong}{Agentic Long-Context Understanding\xspace}
\newcommand{\method}{AgenticLU\xspace}
\newcommand{\coc}{CoC\xspace}
\newcommand{\coclong}{Chain-of-Clarifications\xspace}

\title{Self-Taught Agentic Long-Context Understanding}

\author{
 \textbf{Yufan Zhuang\textsuperscript{1,2}},
 \textbf{Xiaodong Yu\textsuperscript{1}},
 \textbf{Jialian Wu\textsuperscript{1}},
 \textbf{Ximeng Sun\textsuperscript{1}},
 \textbf{Ze Wang\textsuperscript{1}},\\
 \textbf{Jiang Liu\textsuperscript{1}},
 \textbf{Yusheng Su\textsuperscript{1}},
 \textbf{Jingbo Shang\textsuperscript{2}},
 \textbf{Zicheng Liu\textsuperscript{1}},
 \textbf{Emad Barsoum\textsuperscript{1}}
\\
 \textsuperscript{1}AMD,
 \textsuperscript{2}UC San Diego
}

\begin{document}
\maketitle
\begin{abstract}


Answering complex, long-context questions remains a major challenge for large language models (LLMs) as it requires effective question clarifications and context retrieval. 
We propose \methodlong (\method), a framework designed to enhance an LLM’s understanding of such queries by integrating targeted self-clarification with contextual grounding within an agentic workflow.
At the core of \method is Chain-of-Clarifications (\coc), where models refine their understanding through self-generated clarification questions and corresponding contextual groundings. 
By scaling inference as a tree search where each node represents a \coc step, we achieve 97.8\% answer recall on NarrativeQA with a search depth of up to three and a branching factor of eight.
To amortize the high cost of this search process to training, we leverage the preference pairs for each step obtained by the \coc workflow and perform two-stage model finetuning: (1) supervised finetuning to learn effective decomposition strategies, and (2) direct preference optimization to enhance reasoning quality. This enables \method models to generate clarifications and retrieve relevant context effectively and efficiently in a single inference pass.
Extensive experiments across seven long-context tasks demonstrate that \method significantly outperforms state-of-the-art prompting methods and specialized long-context LLMs, achieving robust multi-hop reasoning while sustaining consistent performance as context length grows. \renewcommand\thefootnote{}\footnotetext{Code and data is available at: \url{https://github.com/EvanZhuang/AgenticLU}.}\renewcommand\thefootnote{\arabic{footnote}}

\end{abstract}

\section{Introduction}

\begin{figure*}[t!]
    \begin{center}
        \includegraphics[width=\linewidth, keepaspectratio]{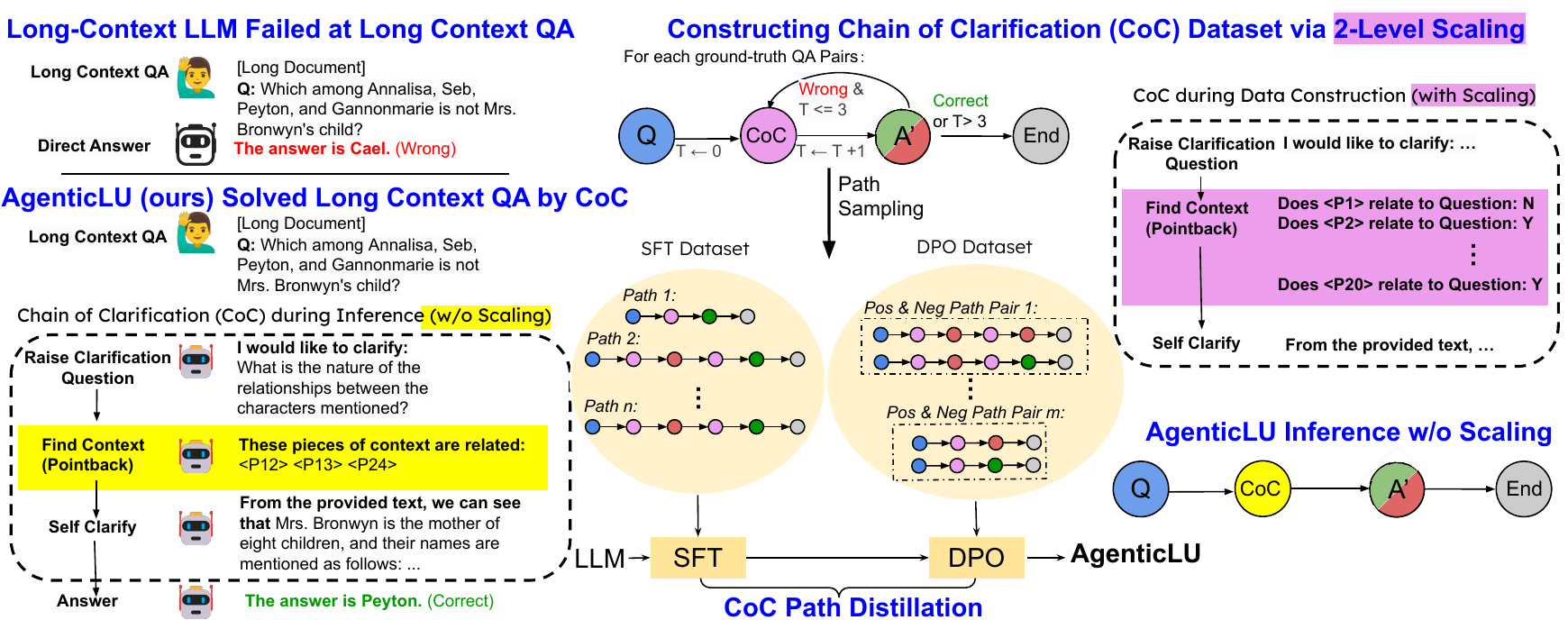}
    \end{center}
    \caption{\textbf{Overview of the AgenticLU pipeline}: The model iteratively refines its understanding of long-context inputs through an agentic workflow. At each step, it raises self-clarifications, retrieves relevant context via the pointback mechanism, and updates its reasoning trace. The framework integrates CoC Path Construction to generate diverse reasoning paths, followed by two-stage fine-tuning (SFT and DPO) to enhance long-context understanding.}
    \label{fig:pipeline}
\end{figure*}

Large language models have achieved notable milestones in natural language processing, demonstrating exceptional performance in tasks such as mathematical reasoning, code generation, and conversational understanding~\cite{openai2023gpt4, deepseekai2025deepseekr1incentivizingreasoningcapability}. 
However, effectively comprehending and utilizing long-context inputs remains a major challenge. Complex queries often require models to retrieve multiple relevant pieces of information from extensive contexts and synthesize them coherently. While recent advancements have extended context windows to 128K and even 2M tokens~\cite{dubey2024llama3, touvron2023llama2, reid2024gemini}, these models still struggle to fully integrate and reason over large-scale contextual information. 
Recent studies~\cite{liu2024lost, prolong} highlight a fundamental challenge in long-context understanding: the disparity between a model’s nominal context size—the theoretical maximum input length—and its effective context window, the portion of the input the model actively utilizes for reasoning. This gap significantly impacts the understanding performance, limiting the model's ability to fully comprehend and integrate long-context information. 

We introduce a novel framework \method to enhance long-context comprehension in LLMs. 
As illustrated in~\cref{fig:pipeline}, the core of \method is \textbf{Chain-of-Clarifications} (\coc), a process where models enhance their understanding by generating clarification questions, retrieving relevant information from the long context and answering their own clarification questions based on the gathered evidence. Rather than relying on a direct response, \coc helps models refine their reasoning iteratively, resolving uncertainties along the way. We structure the framework into the following two stages.

\paragraph{\coc Path Construction.} To collect reliable \coc~understanding path, we structure data collection as a tree search, where each \coc~step represents a node. We leverage extended inference time to determine the effective clarification questions to ask and the relevant evidence to retrieve. With a search depth of three and a branching factor of eight, \method successfully retrieves 97.8\% of the correct answers in NarrativeQA~\cite{kocisky-etal-2018-narrativeqa}, demonstrating its capability to tackle complex questions that require multi-step reasoning over long-context inputs.

\paragraph{\coc Path Distillation.} Once the dataset is collected from the tree-search process, we train the model to generate effective clarifications and contextual groundings in a single pass, eliminating the need for scaling at inference time. This is achieved by distilling these collected paths into LLMs through supervised finetuning (SFT) and direct preference optimization (DPO)~\cite{rafailov2024direct}, effectively amortizing the computational cost from inference to training.






Our method \method~significantly improves model's long-context understanding capabilities without relying on laborious human annotations or stronger teacher models for data generation. Instead, the base model's self-generated CoC paths enables it to teach itself to process long-context inputs more effectively. This approach harnesses the model’s inherent long-context capabilities—previously only accessible through an additional LLM agent—allowing it to independently refine its reasoning and retrieval processes.
Empirically, we demonstrate that \method~consistently boosts performance across a set of question-answering tasks up to 128K tokens, outperforming both prompting-based approaches and other long-context-finetuned LLMs.
By integrating self-clarification and context grounding in an agentic manner, we take a step further toward enabling LLMs to comprehend long contexts.

\section{Related Work}
\label{sec:related_work}
\paragraph{Challenges in Long Context Understanding}
LLMs struggle with long contexts despite supporting up to 2M tokens~\cite{dubey2024llama3,reid2024gemini}. 
The ``lost-in-the-middle'' effect~\cite{liu2024lost} and degraded performance on long-range tasks~\cite{li2023loogle} highlight these issues. To address this, ProLong~\cite{prolong} finetunes base models on a large, carefully curated long-context corpus. While this approach improves performance on long-range tasks, it comes at a significant cost, requiring training with an additional 40B tokens and long-input sequences.



\paragraph{Inference-time Scaling for Long-Context}
The Self-Taught Reasoner (STaR) framework \citep{zelikman2022star} iteratively generates rationales to refine reasoning, with models evaluating answers and finetuning on correct reasoning paths. \citet{wang2024multi} introduced Model-induced Process Supervision (MiPS), automating verifier training by generating multiple completions and assessing accuracy, boosting PaLM 2's performance on math and coding tasks. \citet{li2024large} proposed an inference scaling pipeline for long-context tasks using Bayes Risk-based sampling and fine-tuning, though their evaluation is limited to shorter contexts (10K tokens) compared to ours (128K tokens).

\paragraph{Agentic Workflow for Long-Context} 
Agentic workflows~\cite{yao2022react} enable LLMs to autonomously manage tasks by generating internal plans and refining outputs iteratively. 
The LongRAG framework~\cite{zhao-etal-2024-dual} enables an LLM and an RAG module to collaborate on long-context tasks by breaking down the input into smaller segments, processing them individually, and integrating the results to form a coherent output.
Chain-of-Agents (CoA)~\cite{zhang2024chain} tackles long-context tasks through decomposition and multi-agent collaboration. In CoA, the input text is divided into segments, each handled by a worker agent that processes its assigned portion and communicates its findings to the next agent in the sequence.
Unlike these, our approach employs a single LLM that orchestrates its own reasoning and retrieval without relying on multiple components. By dynamically structuring its process and iteratively refining long-context information, our model reduces complexity while maintaining efficiency.

\section{The Context Size Gap}

State-of-the-art LLMs have made strong claims about their context lengths, supporting hundreds of thousands of input tokens. However, recent studies~\cite{prolong,yen2024helmet,shang2024ai} have shown that the \emph{effective} context size of an LLM (the length over which it can reliably perform tasks such as information retrieval and complex reasoning) often diverges from its claimed, or \emph{nominal}, context length. 

\begin{figure}[t!]
    \begin{center}
    \includegraphics[width=0.8\columnwidth , keepaspectratio]{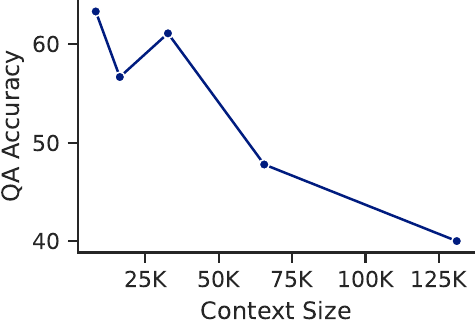}
    \end{center}
    \caption{\textbf{Effective context size is smaller than nominal context size.} 
    Performance of Llama3.1-8B-Instruct (advertised 128K-token context) on the HotPotQA dataset 
    drops sharply as input length increases (8K, 16K, 32K, 64K, 128K), illustrating the 
    gap between nominal and effective context capacities.}
    \label{fig:hotpotQA}
\end{figure}

To illustrate this gap, we evaluate Llama3.1-8B-Instruct, which supports a 128K-token context, on the HotPotQA dataset to test multi-hop QA performance at various input lengths (8K, 16K, 32K, 64K, and 128K). We artificially expand the input by adding irrelevant context and measure the accuracy of its answers  using GPT-4o as a judge. As shown in \cref{fig:hotpotQA}, The model’s performance degrades substantially as increasing context length, demonstrating the discrepancy between nominal and effective context sizes.

While expanding nominal context capacity is undoubtedly important, we argue that it is not sufficient for solving all long-context problems. By analogy with computer memory, simply having more capacity does not guarantee efficient or accurate computation; one must also manage the ``loading'' of relevant information in and out of this memory. Therefore, we propose an agentic workflow aimed at helping LLMs process and interpret extended contexts more intelligently.

\section{\coclong Workflow}
\label{sec:methodology}
Our approach centers on enhancing long-context comprehension through an iterative, self-refining process that blends inference-time scaling with agentic reasoning. 
We coin this agentic workflow Chain-of-Clarifications (\coc).
In this section, we detail its key components, including the self-clarification process and the pointback mechanism, as illustrated in~\cref{fig:pipeline}.

Our proposed \coc framework is designed to mitigate the gap between nominal and effective context sizes in large language models. 
Rather than processing the entire long context and potentially multi-hop questions in a single pass, our methodology decomposes the task into a sequence of targeted sub-tasks. At each \coc step, the model autonomously:

\begin{itemize}
    \item \textbf{Generates clarifying questions} by identifying areas of the long input that require further elaboration or are prone to misinterpretation.
    \item \textbf{Pointbacks to relevant context} by using a pointback mechanism that highlights critical segments of the context by naming the index of relevant paragraphs. 
    In the data collection phase, this is done by iteratively querying the LLM about the relevance of each paragraph with respect to the question.
    After training, the model is finetuned to generate the related paragraph indexes directly in a single pass.  
    \item \textbf{Answers clarifying questions} by integrating highlighted context into consideration to build a more accurate and contextually grounded understanding of the long document.
    \item \textbf{Answers the original question} by combining all newly gathered clarifications, the model attempts to generate a valid answer to the original question.
\end{itemize}

It is important to note a key distinction between \coc path generation during data collection and the actual task deployment of the agentic workflow. In the data generation phase, we prompt the LLM to iteratively process each chunk of input text along with its self-generated clarifying questions, ensuring accurate retrieval of relevant context. 
During training, rather than relying on repeated inference calls, we finetune the model to directly generate the indexes of relevant paragraphs using pointback examples, effectively amortizing the computational cost into training. This enables the model to internalize the retrieval process, allowing it to dynamically synthesize relevant clarifications and contextual references at inference time without requiring extensive additional prompting.

\section{Data Generation \& Model Training}

\paragraph{Dataset} 
We use the NarrativeQA~\cite{kocisky-etal-2018-narrativeqa} dataset to facilitate long-context QA and generate agentic workflow traces with 14.7K QA pairs in the training set. NarrativeQA is designed for reading comprehension over narrative texts, such as books and movie scripts, where each example includes a full story and a set of corresponding QA pairs. This dataset emphasizes deeper reasoning and long-context understanding, as many questions require synthesizing information from multiple parts of the narrative rather than focusing solely on particular local context. Its relatively long passages make NarrativeQA particularly suitable for testing and refining agentic reasoning in large language models, as the answers often depend on weaving together details spanning the entire text.

\paragraph{Base Model} Our base model is \textit{Llama3.1-8B-Instruct}~\cite{dubey2024llama3}, an 8-billion-parameter instruction-tuned Llama model. This model is built on the same transformer architecture as Llama3, but with additional fine-tuning data to improve its performance on multi-turn dialogue and instruction-following tasks.

\subsection{\coc Path Construction}
We employ a test-time scaling approach to generate \coc paths. For each question, we construct a tree of search paths where each node represents a distinct clarification question posed by the LLM.

In our experiments, we use a branching factor of 8 at each depth and select the most promising trace based on an evaluation score that combines:
\begin{itemize}
    \item \textbf{Semantic similarity}, measured by the RougeL~\cite{lin-2004-rouge} score relative to the ground truth.
    \item \textbf{Discrete correctness}, evaluated by a binary verification using GPT4o-mini.
\end{itemize}

In the data construction process, the relevant context is found by iteratively querying the LLM about the relevance of all chunked passages. Here we use 512 as the chunk size. This process is compute-intensive but only happens in data collection. 
After the training, the LLM will directly generate the paragraph numbers of the relevant context as shown in the lower right of~\cref{fig:pipeline}.

For most long-context tasks, a single clarification question suffices because the required reasoning is not highly complex. 92\% of the questions in our experiments are resolved correctly with just one round of clarification. More challenging tasks may require multiple rounds of clarification: two rounds resolve 53\% of the remaining 8\%, and three rounds resolve 35\% of the remaining 4\%. Because of the exponentially increasing cost—and given that 97.4\% of the training questions are already solved—we limit the maximum depth of our inference scaling to 3.

The statistics of the collected dataset are shown in~\cref{tab:narrativeqa_stats}. The total number of conditional generation tokens that the LLM trained on is 17M tokens, with input that has an average length of 67K and a max length of 128K tokens.

\begin{table}[t]
\centering

\caption{Statistics of the generated traces dataset used in finetuning derived from NarrativeQA. We left out 11.9K traces for validation.}
\resizebox{0.75\columnwidth}{!}{
\begin{tabular}{l r}
\toprule
 \textbf{Data} & \# \\
\midrule
Num of Traces             & 107,550  \\
Avg Context Length            & 67,812  \\
Avg Chosen Response Length & 165  \\
Avg Rejected Response Length & 164  \\
Total Generation Tokens    & 17M  \\
\bottomrule
\end{tabular}}
\label{tab:narrativeqa_stats}
\end{table}

\subsection{\coc Path Distillation}
We employ a two-stage finetuning recipe: Supervised Fine-Tuning (SFT) followed by Direct Preference Optimization (DPO)~\cite{rafailov2024direct}, to convert our base model into a long-context understanding agent.
The dataset statistics is described in~\cref{tab:narrativeqa_stats}, with input length up to 128K tokens. 

\paragraph{Supervised Fine-Tuning} In the first phase, we finetune \textit{Llama3.1-8B-Instruct} using the generated \coc paths. Each training example includes (1) the full context from NarrativeQA, (2) the question, and (3) the step-by-step reasoning trace leading to the final answer. 
By exposing the model to these traces, we encourage it to internalize multi-step reasoning strategies and context grounding for the long-context inputs. 
The SFT stage uses a standard cross-entropy loss on the next-token prediction task, ensuring the model learns how to produce consistent and complete reasoning sequences.

\paragraph{Direct Preference Optimization} 
In the second phase, we apply Direct Preference Optimization to further refine the model’s output quality. 
To create preference pairs, we sample incorrect workflow traces as negative examples with using GPT4o-mini as the judge for answer correctness from the test-time scaling. 
DPO explicitly optimizes the model to generate higher-ranked responses more frequently, thus aligning the agent’s outputs with desirable characteristics, such as clarity, correctness, and coherence. This stage ensures that even among valid reasoning paths, the model learns to prioritize the most instructive reasoning.

The details for the two-phase training are listed in~\cref{asec:hyperparameters}.

\section{Evaluation}
\label{sec:evaluation}

\begin{table*}[ht]
  \centering
  \caption{Performance difference of \method and its base, Llama3.1-8B-Instruct ($\delta=$\method-8B minus Llama3.1-8B), on long context (the 128K tasks) and short-context benchmarks (6 regular tasks including ARC, GSM8K, and MMLU), the details of the short-context performance can be found in~\cref{asec:short_context}. Scores represent accuracy, with \method demonstrating significantly improved performance across long-context tasks with minimal effect on regular task performance.}
  \label{tab:performance-long-benchmarks}
  \resizebox{\textwidth}{!}{
\begin{tabular}{l|r|rrrrrrrr}
  \toprule
  \textbf{Model} & \textbf{Short Avg} & HotpotQA & Natural Questions & TriviaQA & PopQA & NarrativeQA & InfiniQA & InfiniChoice & \textbf{Long Avg}  \\
  \midrule
  Llama3.1-8B  & \textbf{62.3}  & 40.0   & 56.1    & 80.6    & 56.1   & 38.0    & 48.0   & 55.0     & \textbf{53.4}  \\
 AgenticLU ($\delta$)    & \textbf{-0.6} & +31.1 & +21.7 & +7.7 & +9.4 & +18.0  & +2.0 & +13.0  & \color{red}{\textbf{+14.7}} \\
  \bottomrule
\end{tabular}
}
\end{table*}

\begin{figure*}[t!]
    \begin{center}
    \includegraphics[width=\linewidth , keepaspectratio]{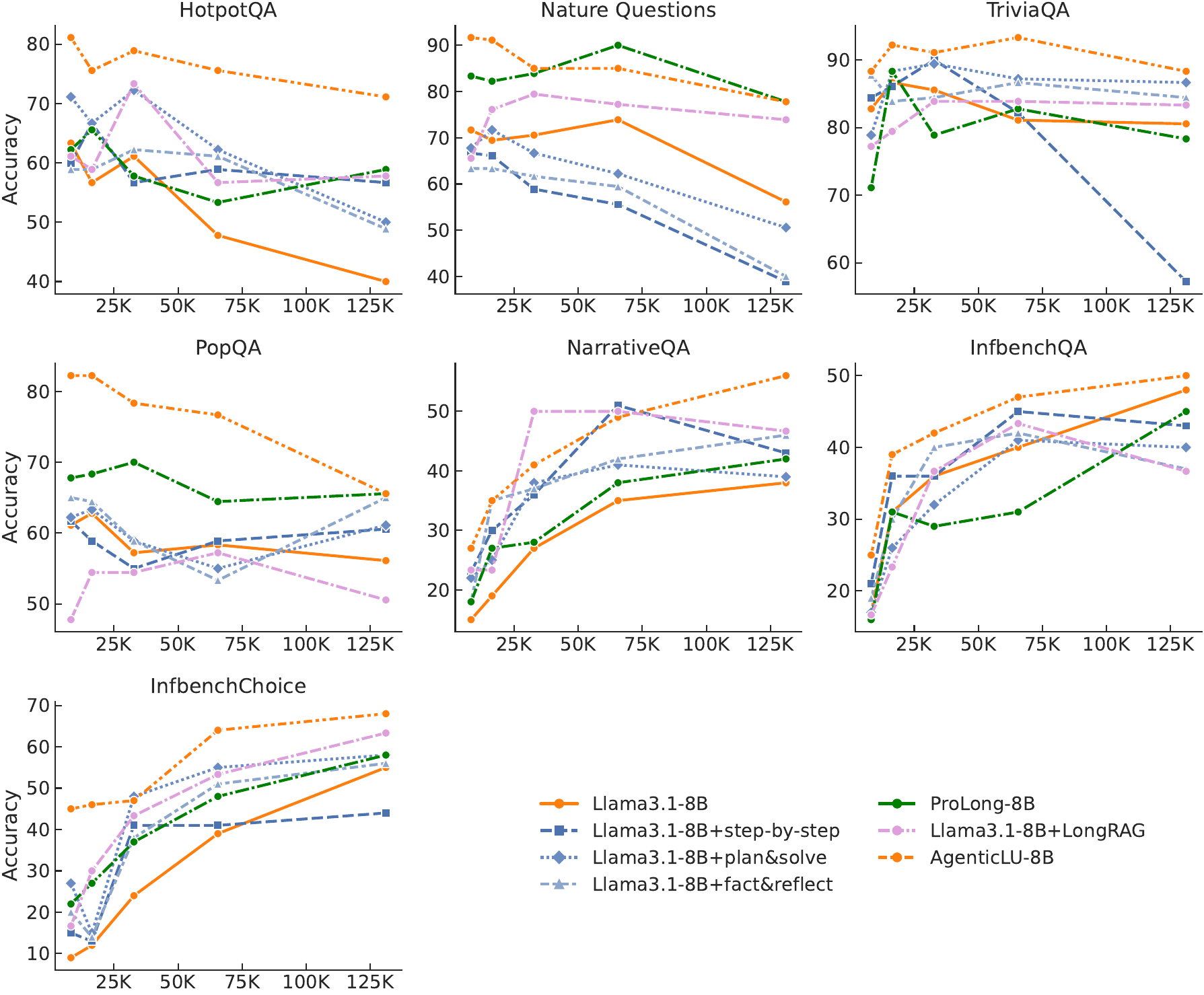}
    \end{center}
    \caption{\textbf{Main results on 7 long-context tasks across context lengths from 8K to 128K.} Our \method-8B (dotted orange) achieves significant improvements on \emph{all} tasks over our base model Llama3.1-8B (solid orange). We also compare with the prompting methods (Step-by-Step, Plan-and-Solve, Fact-and-Reflect, LongRAG) and the state-of-the-art ProLong-8B model. \method-8B consistently maintains strong performance across most tasks and context lengths.}
    \label{fig:rag_result}
\end{figure*}

In this section, we assess our method~\method using a suite of evaluation tasks drawn from the HELMET long-context benchmark~\cite{yen2024helmet}. Our experiments focus on testing models’ ability to retain, process, and reason over extended contexts ranging from 8K to 128K tokens.

\subsection{Tasks and Metrics}
We evaluate our models and baselines on the Helmet~\cite{yen2024helmet} long-context evaluation benchmark's retrieval-augmented generation (RAG) and long-range QA (LongQA) tasks ranging from 8K, 16K, 32K, 64K, to 128K.

We use GPT-4o as the judge for answer correctness, with the prompt template shown in~\cref{asec:prompt_template}. 
We report accuracies for all datasets.

The RAG test suite includes: 
(1) \textbf{HotpotQA}~\cite{yang-etal-2018-hotpotqa}, a multi-hop reasoning dataset over Wikipedia; 
(2) \textbf{Natural Questions}~\cite{kwiatkowski2019natural}, real user queries with Wikipedia-based short and long answers; 
(3) \textbf{TriviaQA}~\cite{JoshiTriviaQA2017}, a large-scale trivia dataset with question-answer pairs linked to evidence documents; 
(4) \textbf{PopQA}~\cite{mallen2023llm_memorization}, a dataset testing model memorization with fact-based questions from popular culture.

The LongQA test suite includes: 
(1) \textbf{NarrativeQA}~\cite{kocisky-etal-2018-narrativeqa}, a reading comprehension dataset with Wikipedia summaries and story-based Q\&A; 
(2) \textbf{InfiniteBench QA}~\cite{zhang-etal-2024-bench}, a long-range QA benchmark requiring reasoning over extended contexts; 
(3) \textbf{InfiniteBench Multiple-Choice}~\cite{zhang-etal-2024-bench}, a multiple-choice variant of the previous evaluating reading comprehension over long documents.

For the four RAG tasks, each question is put alongside a set of relevant contexts, and the overall input length is increased by appending irrelevant context. Consequently, these tasks become strictly more difficult as the context window expands. 
In contrast, for the three LongQA tasks, the relevant context may not appear in the truncated input (the first 8K, 16K, or 128K tokens). Hence, performance might improve at longer input lengths simply because the necessary information becomes available only after including more tokens.


\subsection{Baselines} 

We compare~\method against a diverse set of strong baselines representing different approaches for handling long-context tasks. Our comparisons include two main categories. 

Under prompting methods we consider techniques that require no additional model training. In particular, we evaluate (a) the chain-of-thought approach~\cite{kojima2022large}, which encourages models to decompose complex questions into intermediate reasoning steps; (b) fact-and-reflection prompting~\cite{zhao-etal-2024-fact}, which iteratively verifies and refines factual claims to enhance consistency; (c) plan-and-solve prompting~\cite{wang2023plan}, where the model first outlines a high-level plan before sequentially executing it to address structured reasoning tasks; and (d) LongRAG~\cite{zhao-etal-2024-longrag} where a hybrid RAG system is used to retrieve relevant context to generate global summaries and local details~\footnote{Note that LongRAG provided finetuned models as well. But the SFT-ed Llama3-8B only supports 8K context length. Thus we did not include it in our comparison.}. 

In the fine-tuning category, we focus on models that have been specifically adapted for extended context data. For a substantial comparison, we employ Prolong-8B-512K~\cite{prolong}—a model based on the Llama3 8B architecture that has been further trained on an additional 40B tokens of long-context data.

\subsection{Main Results}

The performance of \method and baseline models is shown in~\cref{fig:rag_result}. 


\paragraph{Self-clarification significantly improves multi-hop reasoning.} \method-8B consistently surpasses other methods in HotpotQA. By iteratively refining its understanding, resolving ambiguities, and verifying intermediate steps, the model achieves higher accuracy, particularly as context length increases.

\paragraph{Robust performance across diverse datasets.} Unlike baseline models, \method-8B maintains consistently strong performance across RAG and LongQA benchmarks, demonstrating its ability to adapt effectively to different long-context tasks.

\paragraph{Reduced performance degradation with longer contexts.} While most models experience significant accuracy drops as context length increases, \method-8B remains stable. Its self-clarification and pointback mechanisms effectively filter noise from irrelevant information, allowing the model to extract and prioritize essential evidence.

\paragraph{Fine-tuning vs. prompting trade-offs.} While structured prompting techniques like \textit{plan-and-solve} improve short-context reasoning, they struggle with extreme context lengths (e.g., 128K tokens). In contrast, \method-8B, through targeted finetuning with self-clarification and pointback, maintains robust long-context reasoning without relying on complex prompting strategies. Although ProLong-8B, another finetuned model, achieves strong results, it comes with significantly higher training costs. \method-8B, by contrast, is more data-efficient and generalizes better to novel tasks, making it a more practical and effective solution for long-context reasoning.

Overall, these results underscore the effectiveness of \method-8B in tackling long-context understanding challenges. The integration of self-clarification plays a crucial role in improving grounding, reasoning, and comprehension in long-context settings.

\subsection{Performance on Short-Context Tasks}
To demonstrate that our fine-tuning process preserves the model's general capabilities while enhancing long-context understanding, we evaluated the finetuned model on a diverse set of standard benchmarks. These include elementary and advanced reasoning tasks ARC Easy and ARC Challenge~\cite{Clark2018ThinkYH}, mathematical problem-solving GSM8K~\cite{cobbe2021training}, MathQA~\cite{amini2019mathqa}, and broad knowledge assessment MMLU~\cite{hendryckstest2021, hendrycks2021ethics}, MMLU-Pro~\cite{wang2024mmlupro}.

We report the average performance across short-context tasks in~\cref{tab:performance-long-benchmarks}, and each individual task result can be found in~\cref{asec:short_context}. We find that the short-context performance is well preserved, demonstrating that \method's core reasoning and problem-solving abilities remain strong and are not compromised by the significant improvements to its long-context understanding powers.


\section{Analyses \& Ablation Studies}
\label{sec:analysis}
In this section, we take a closer look at how each part of our approach affects long-context understanding and retrieval. Specifically, we study three main questions: (1) Can the finetuned system benefit from multi-round \coc? (2) Does adding clarifications and pointing back to the original document help the model understand and utilize the context more accurately? (3) How much additional compute overhead does \method add to the process?

\begin{table}
\centering
\caption{We evaluate the performance of adding additional self-clarification and contextual grounding rounds at inference time. The gain from self-clarification is close to optimal at the initial round.}
\resizebox{\columnwidth}{!}{
\begin{tabular}{lccccc}
\toprule
\textbf{Model} & \textbf{HotpotQA} & \textbf{NaturalQ} & \textbf{PopQA} & \textbf{TriviaQA} & \textbf{Avg} \\
\midrule
Llama-3.1-8B         & 40.0  & 56.1  & 56.1  & 80.6  & 58.2 \\
\method-8B           & 71.1  & 77.8  & 65.5  & 88.3  & 75.7 \\
\quad (w/ 2 rounds)  & 71.1  & 76.7  & 67.2  & 91.7  & 76.7 \\
\quad (w/ 3 rounds)  & 75.5  & 78.8  & 68.3  & 91.1  & 78.4 \\
\bottomrule
\end{tabular}
}
\label{tab:multiround}
\end{table}

\subsection{How many rounds of \coc are needed?}
\paragraph{Setup.}
We add additional rounds of reasoning in the evaluation and see if the LLM can benefit from multi-rounds of reasoning at test-time.

\paragraph{Analysis.}
The results, presented in Table~\ref{tab:multiround}, indicate that additional rounds of agentic reasoning do provide performance improvements. 

This suggests that while significant benefits of self-clarification are achieved in the first round, additional rounds still contribute to further improvements. 
One possible explanation is the nature of our dataset: approximately 92\% of the questions are resolved within a single round of clarification. However, for the remaining cases, extended reasoning allows the model to refine its understanding, leading to measurable gains in performance with more clarification and reasoning.


\begin{table}
\centering

\caption{We test the agentic workflow with \method-8B when taking out the self-clarification steps and the contextual grounding (pointback) step. The tasks are with 128K context length.}
\resizebox{\columnwidth}{!}{
\begin{tabular}{lccccc}
\toprule
\textbf{Model} & \textbf{HotpotQA} & \textbf{NaturalQ} & \textbf{PopQA} & \textbf{TriviaQA} & \textbf{Avg} \\
\midrule
Llama-3.1-8B               & 40.0  & 56.1  & 56.1  & 80.6  & 58.2 \\
\method-8B                 & 71.1  & 77.8  & 65.5  & 88.3  & 75.7 \\
\quad (w/o Clarification)  & 57.8  & 56.7  & 55.5  & 78.3  & 62.1 \\
\quad (w/o Pointback)      & 53.3  & 59.4  & 52.7  & 83.3  & 62.2 \\
\bottomrule
\end{tabular}
}
\label{tab:ablation}
\end{table}
\subsection{Do Self-Clarifications and Pointback Help in Long-Context Understanding?}
\paragraph{Setup.}
To evaluate the impact of each component in our agentic workflow, we compare the full \method-8B model against two variants: one without the self-clarification step and another without the contextual grounding (\emph{pointback}) step. 
We use the four RAG datasets with 128K context length as the evaluation benchmark, and compare the performance alongside the original model.

\paragraph{Analysis.}
Table~\ref{tab:ablation} shows the results on four QA benchmarks with a 128K context length. Removing self-clarification leads to an absolute performance drop of at least 10 points across most tasks (e.g., from 71.1\% to 57.8\% on HotpotQA), confirming that the model benefits from clarifying its own uncertainties when the context is long. Meanwhile, omitting pointback yields degenerate results, indicating that pinpointing relevant information at each stage is crucial for long-context QA. Overall, these findings highlight the importance of both clarifications and context-grounding to maximize retrieval accuracy and robustness in lengthy documents.

\subsection{How much additional compute cost does \method impose in generation?}
Since additional generation steps are introduced in the QA process, we assess the overhead in inference time.
Naïvely, long-context inference and multi-round conversations could significantly amplify compute costs. However, by leveraging prefix caching to store computed KV caches, the additional cost scales linearly with the number of newly generated tokens rather than exponentially.

To quantify this overhead, we conduct a runtime evaluation on 100 queries with a 128K context size. The results, summarized in~\cref{tab:runtime_overhead}, demonstrate that the additional computational overhead remains minimal when using prefix caching.
\begin{table}
  \centering
  \caption{Performance Overhead Comparison between direct answering baseline and \method.}
  \label{tab:runtime_overhead}
  \resizebox{\columnwidth}{!}{
  \begin{tabular}{lcc}
    \toprule
    \textbf{Metric} & \textbf{Baseline} & \textbf{\method} \\
    \midrule
    Runtime Overhead & 100\%   & 101.93\% \\
    Avg Tokens Generated in One Round  & 76.28   & 1205.38    \\
    \bottomrule
  \end{tabular}
}
\end{table}

\section{Conclusion}
\label{sec:conclusion}

In this work, we introduce Agentic Long-Context Understanding (\method), a framework designed to enhance large language models' ability to process and reason over long-context inputs with self-generated data. 
By incorporating an agentic workflow (\coc) that dynamically refines model reasoning through self-clarifications and contextual grounding, \method significantly improves LLM's long context understanding capabilities.

Through a combination of trace data collection and two-stage post-training, our approach enables models to autonomously explore multiple reasoning paths, distill the most effective clarification strategies, and improve their understanding of lengthy documents. 
Extensive evaluations on long-context benchmarks demonstrate that AgenticLU outperforms existing prompting techniques and finetuned baselines, maintaining strong performance across context lengths up to 128K tokens. 
Additionally, ablation studies confirm that self-clarification and pointback mechanisms play a crucial role in improving retrieval and reasoning over long-contexts.

\section*{Limitations}
Despite its effectiveness in long-context reasoning, \method has notable limitations. One key drawback is its inability to autonomously determine when to stop multi-round reasoning. While additional rounds of self-clarification can improve performance, the model follows a fixed number of reasoning steps rather than dynamically assessing when further refinement is necessary. This can lead to inefficiencies, where the model either stops too early, missing potential improvements, or continues reasoning unnecessarily, expending computational resources without significant gains.

Developing a fully agentic mechanism remains an open challenge. Ideally, the model should assess its confidence in an intermediate response and decide whether further clarification is needed. Future work should explore approaches that enable AgenticLU to regulate its reasoning depth dynamically, optimizing both efficiency and performance.

\section*{Acknowledgement}
Our work is sponsored in part by NSF CAREER Award 2239440, NSF Proto-OKN Award 2333790, Sponsored Research Projects from companies like Cisco and eBay, as well as generous gifts from Google, Adobe, and Teradata. Any opinions, findings, and conclusions or recommendations expressed herein are those of the authors and should not be interpreted as necessarily representing the views, either expressed or implied, of the U.S. Government. The U.S. Government is authorized to reproduce and distribute reprints for government purposes not withstanding any copyright annotation hereon.


\bibliography{anthology,custom}

\appendix

\section{Training Configurations}
\label{asec:hyperparameters}
We employed the DeepSpeed~\cite{rasley2020deepspeed} framework for distributed training across four GPU nodes, each equipped with four AMD MI250 GPUs. 
We used vLLM~\cite{kwon2023efficient} for inference.
Our implementation builds upon OpenRLHF~\cite{hu2024openrlhf} for both SFT and DPO. 
Given the input sequence length of up to 128K tokens, we leveraged FlashAttention-2~\cite{dao2023flashattention} alongside Ring Attention~\cite{liu2023ring} to efficiently process extremely long sequences. 
The detailed hyperparameters for SFT and DPO are provided in \cref{tab:hyperparameters_sft} and \cref{tab:hyperparameters_dpo}.
\begin{table}[h]
    \centering
    \caption{Hyperparameters for SFT.}
    \small
    \begin{tabular}{l|l}
        \toprule
        \textbf{Hyperparameter} & \textbf{Value} \\
        \midrule
        Learning Rate & 5e-7 \\
        Learning Rate Schedule & Cosine Annealing \\
        Optimizer & Adam \\
        $\beta_1$ & 0.9 \\
        $\beta_2$ & 0.95 \\
        Training dtype & bf16 \\
        Batch Size & 128 \\
        Max Length & 131,072 \\
        \bottomrule
    \end{tabular}
    \label{tab:hyperparameters_sft}
\end{table}
\begin{table}[h]
    \centering
    \caption{Hyperparameters for DPO.}
    \small
    \begin{tabular}{l|l}
        \toprule
        \textbf{Hyperparameter} & \textbf{Value} \\
        \midrule
        Learning Rate & 5e-7 \\
        Learning Rate Schedule & Cosine Annealing \\
        Optimizer & Adam \\
        $\beta_1$ & 0.9 \\
        $\beta_2$ & 0.95 \\
        Training dtype & bf16 \\
        Batch Size & 128 \\
        $\beta$ & 0.1 \\
        Max Length & 131,072 \\
        \bottomrule
    \end{tabular}
    \label{tab:hyperparameters_dpo}
\end{table}

\section{Short Context Performance}
\label{asec:short_context}
\begin{table*}[ht]
  \centering
  \caption{Performance comparison of \method and Llama3.1-8B-Instruct on short-context benchmarks. Scores represent accuracy percentages, with \method demonstrating matching results across tasks.}
  \label{tab:performance-benchmarks}
  \resizebox{\textwidth}{!}{
  \begin{tabular}{lccccccc}
    \toprule
    \textbf{Model} & \textbf{ARC Easy} & \textbf{ARC Challenge} & \textbf{GSM8k} & \textbf{MathQA} & \textbf{MMLU} & \textbf{MMLU Pro} & \textbf{Avg} \\
    \midrule
    Llama3.1-8B     & 84.80      & 59.64          & 80.13  & 42.88  & 68.72 & 37.71 & 62.31     \\
    \method-8B & 83.96 & 58.36 & 80.51 & 41.74 & 68.38 & 37.51 & 61.74 \\    
    \bottomrule
  \end{tabular}
  }
\end{table*}

As shown in~\cref{tab:performance-benchmarks}, we evaluate the short-context perfromance across six tasks: ARC Easy, ARC Challenge, GSM8K, MathQA, MMLU, and MMLU Pro. \method performs on par with the base model Llama3.1-8B-Instruct on short-context benchmarks, demonstrating that \method preserves the original short-context ability while greatly enhancing long-context performance.

\section{Agentic Workflow without Training}
We conducted an additional experiment for prompting only with our agentic workflow, and we find that with an option to generate clarifications the model does get better on multi-hop QA questions (HotPotQA), but it is generally difficult for the base model to point to the correct paragraphs directly without any training, hence often resulting in same or slightly worse performance.

Results are listed in~\cref{tab:prompting_results}, the context length is 128K tokens.

\begin{table*}[h]
\centering
\caption{Performance comparison across different QA benchmarks with 128K token context length, with or without training.}
\begin{tabular}{lcccccc}
\toprule
\textbf{Model} & \textbf{HotpotQA} & \textbf{NaturalQ} & \textbf{PopQA} & \textbf{TriviaQA} & \textbf{Avg} \\
\midrule
Llama-3.1-8B & 40.0 & 56.1 & 56.1 & 80.6 & 58.2 \\
Llama-3.1-8B + Prompting Only & 53.3 & 56.7 & 51.6 & 72.8 & 58.6 \\
AgenticLU-8B & 71.1 & 77.8 & 65.5 & 88.3 & 75.7 \\
\bottomrule
\end{tabular}

\label{tab:prompting_results}
\end{table*}

\section{Detailed Results on Seven Benchmark Tasks}
As shown in~\cref{tab:hotpotqa},~\cref{tab:nature_questions},~\cref{tab:triviaqa},~\cref{tab:popqa},~\cref{tab:narrativeqa},~\cref{tab:infbenchqa} and~\cref{tab:infbenchchoice}, we evaluate the long-context performance across seven tasks: HotpotQA, Natural Questions, TriviaQA, PopQA, NarrativeQA, InfiniteBench QA and InfiniteBench Multiple-Choice. 
\method provides significant improvement for all tasks, especially for those that require multi-hop reasoning such as HotPotQA.

\begin{table}[t]
\centering
\footnotesize
\resizebox{\columnwidth}{!}{
\begin{tabular}{lccccc}
\toprule
 & \multicolumn{5}{c}{HotpotQA} \\
Model & 8K & 16K & 32K & 64K & 128K \\
\midrule
Llama3.1-8B               & 63.3 & 56.7 & 61.1 & 47.8 & 40.0 \\
Llama3.1-8B+step-by-step  & 60.0 & 66.7 & 56.7 & 58.9 & 56.7 \\
Llama3.1-8B+plan\&solve   & 71.1 & 66.7 & 72.2 & 62.2 & 50.0 \\
Llama3.1-8B+fact\&reflect & 58.9 & 58.9 & 62.2 & 61.1 & 48.9 \\
ProLong-8B                & 62.2 & 65.6 & 57.8 & 53.3 & 58.9 \\
Llama3.1-8B+LongRAG       & 61.1 & 58.9 & 73.3 & 56.7 & 57.8 \\
\method-8B                & 81.1 & 75.6 & 78.9 & 75.6 & 71.1 \\
\bottomrule
\end{tabular}}
\caption{Long-context performance on HotpotQA.}
\label{tab:hotpotqa}
\end{table}

\begin{table}[t]
\centering
\footnotesize
\resizebox{\columnwidth}{!}{
\begin{tabular}{lccccc}
\toprule
 & \multicolumn{5}{c}{Nature Questions} \\
Model & 8K & 16K & 32K & 64K & 128K \\
\midrule
Llama3.1-8B               & 71.7 & 69.4 & 70.6 & 73.9 & 56.1 \\
Llama3.1-8B+step-by-step  & 66.7 & 66.1 & 58.9 & 55.6 & 38.9 \\
Llama3.1-8B+plan\&solve   & 67.8 & 71.7 & 66.7 & 62.2 & 50.6 \\
Llama3.1-8B+fact\&reflect & 63.3 & 63.3 & 61.7 & 59.4 & 40.0 \\
ProLong-8B                & 83.3 & 82.2 & 83.9 & 90.0 & 77.8 \\
Llama3.1-8B+LongRAG       & 65.6 & 76.1 & 79.4 & 77.2 & 73.9 \\
\method-8B                & 91.7 & 91.1 & 85.0 & 85.0 & 77.8 \\
\bottomrule
\end{tabular}}
\caption{Long-context performance on Nature Questions.}
\label{tab:nature_questions}
\end{table}

\begin{table}[t]
\centering
\footnotesize
\resizebox{\columnwidth}{!}{
\begin{tabular}{lccccc}
\toprule
 & \multicolumn{5}{c}{TriviaQA} \\
Model & 8K & 16K & 32K & 64K & 128K \\
\midrule
Llama3.1-8B               & 82.8 & 86.7 & 85.6 & 81.1 & 80.6 \\
Llama3.1-8B+step-by-step  & 84.4 & 86.1 & 90.0 & 82.2 & 57.2 \\
Llama3.1-8B+plan\&solve   & 78.9 & 88.3 & 89.4 & 87.2 & 86.7 \\
Llama3.1-8B+fact\&reflect & 87.8 & 83.9 & 84.4 & 86.7 & 84.4 \\
ProLong-8B                & 71.1 & 88.3 & 78.9 & 82.8 & 78.3 \\
Llama3.1-8B+LongRAG       & 77.2 & 79.4 & 83.9 & 83.9 & 83.3 \\
\method-8B                & 88.3 & 92.2 & 91.1 & 93.3 & 88.3 \\
\bottomrule
\end{tabular}}
\caption{Long-context performance on TriviaQA.}
\label{tab:triviaqa}
\end{table}

\begin{table}[t]
\centering
\footnotesize
\resizebox{\columnwidth}{!}{
\begin{tabular}{lccccc}
\toprule
 & \multicolumn{5}{c}{PopQA} \\
Model & 8K & 16K & 32K & 64K & 128K \\
\midrule
Llama3.1-8B               & 61.1 & 62.8 & 57.2 & 58.3 & 56.1 \\
Llama3.1-8B+step-by-step  & 61.7 & 58.9 & 55.0 & 58.9 & 60.6 \\
Llama3.1-8B+plan\&solve   & 62.2 & 63.3 & 58.9 & 55.0 & 61.1 \\
Llama3.1-8B+fact\&reflect & 65.0 & 64.4 & 58.9 & 53.3 & 65.0 \\
ProLong-8B                & 67.8 & 68.3 & 70.0 & 64.4 & 65.6 \\
Llama3.1-8B+LongRAG       & 47.8 & 54.4 & 54.4 & 57.2 & 50.6 \\
\method-8B                & 82.2 & 82.2 & 78.3 & 76.7 & 65.6 \\
\bottomrule
\end{tabular}}
\caption{Long-context performance on PopQA.}
\label{tab:popqa}
\end{table}

\begin{table}[t]
\centering
\footnotesize
\resizebox{\columnwidth}{!}{
\begin{tabular}{lccccc}
\toprule
 & \multicolumn{5}{c}{NarrativeQA} \\
Model & 8K & 16K & 32K & 64K & 128K \\
\midrule
Llama3.1-8B               & 15.0 & 19.0 & 27.0 & 35.0 & 38.0 \\
Llama3.1-8B+step-by-step  & 23.0 & 30.0 & 36.0 & 51.0 & 43.0 \\
Llama3.1-8B+plan\&solve   & 22.0 & 25.0 & 38.0 & 41.0 & 39.0 \\
Llama3.1-8B+fact\&reflect & 18.0 & 35.0 & 37.0 & 42.0 & 46.0 \\
ProLong-8B                & 18.0 & 27.0 & 28.0 & 38.0 & 42.0 \\
Llama3.1-8B+LongRAG       & 23.3 & 23.3 & 50.0 & 50.0 & 46.7 \\
\method-8B                & 27.0 & 35.0 & 41.0 & 49.0 & 56.0 \\
\bottomrule
\end{tabular}}
\caption{Long-context performance on NarrativeQA.}
\label{tab:narrativeqa}
\end{table}

\begin{table}[t]
\centering
\footnotesize
\resizebox{\columnwidth}{!}{
\begin{tabular}{lccccc}
\toprule
 & \multicolumn{5}{c}{InfbenchQA} \\
Model & 8K & 16K & 32K & 64K & 128K \\
\midrule
Llama3.1-8B               & 17.0 & 31.0 & 36.0 & 40.0 & 48.0 \\
Llama3.1-8B+step-by-step  & 21.0 & 36.0 & 36.0 & 45.0 & 43.0 \\
Llama3.1-8B+plan\&solve   & 17.0 & 26.0 & 32.0 & 41.0 & 40.0 \\
Llama3.1-8B+fact\&reflect & 19.0 & 30.0 & 40.0 & 42.0 & 37.0 \\
ProLong-8B                & 16.0 & 31.0 & 29.0 & 31.0 & 45.0 \\
Llama3.1-8B+LongRAG       & 16.7 & 23.3 & 36.7 & 43.3 & 36.7 \\
\method-8B                & 25.0 & 39.0 & 42.0 & 47.0 & 50.0 \\
\bottomrule
\end{tabular}}
\caption{Long-context performance on InfbenchQA.}
\label{tab:infbenchqa}
\end{table}

\begin{table}[t]
\centering
\footnotesize
\resizebox{\columnwidth}{!}{
\begin{tabular}{lccccc}
\toprule
 & \multicolumn{5}{c}{InfbenchChoice} \\
Model & 8K & 16K & 32K & 64K & 128K \\
\midrule
Llama3.1-8B               & 9.0  & 12.0 & 24.0 & 39.0 & 55.0 \\
Llama3.1-8B+step-by-step  & 15.0 & 13.0 & 41.0 & 41.0 & 44.0 \\
Llama3.1-8B+plan\&solve   & 27.0 & 15.0 & 48.0 & 55.0 & 58.0 \\
Llama3.1-8B+fact\&reflect & 20.0 & 14.0 & 38.0 & 51.0 & 56.0 \\
ProLong-8B                & 22.0 & 27.0 & 37.0 & 48.0 & 58.0 \\
Llama3.1-8B+LongRAG       & 16.7 & 30.0 & 43.3 & 53.3 & 63.3 \\
\method-8B                & 45.0 & 46.0 & 47.0 & 64.0 & 68.0 \\
\bottomrule
\end{tabular}}
\caption{Long-context performance on InfbenchChoice.}
\label{tab:infbenchchoice}
\end{table}

\section{\coclong Workflow}
The input was first processed into chunks and grouped with paragraph tags.
We list the example prompts used in~\method workflow below. In training, we sampled 100 variations of the same prompt text and use them randomly to avoid training collapse.

\begin{tcolorbox}[
    colback=black!5, colframe=black!50, 
    fontupper=\ttfamily\small, 
    title=\coclong Workflow Prompts
]
\textbf{[System Prompt]}

You are an AI assistant specialized in long context reasoning. Analyze information thoroughly while maintaining clarity and focus. Track the full context of conversations, building connections between concepts and flagging when context review is needed. Break down complex problems into components, showing your reasoning steps and stating key assumptions. Structure your responses with clear headers and periodic summaries. Present evidence for your conclusions, acknowledge uncertainties, and request clarification when needed. Keep your analysis organized, explicit, and focused on addressing the core question.

\textbf{[Long-Context Input]}

<para 1> [chunk 1] </para 1> <para 2> [chunk 2] </para 2> ...
\{Question\}

\textbf{[Self Clarification - Raise Question]}

In order to answer this question, ask one question about what you want to know in order to better answer it.

\textbf{[Contextual Grounding - Pointback]}

Help me find relevant context to answer the previous clarifying question.

\textbf{[Self Clarification - Answer Question]}

Based on the relevant context, answer the previous clarifying question.

\textbf{[Answer the Original Question]}

Now, let's answer the final question. Be concise in your answer.
\end{tcolorbox}

\clearpage

\section{Evaluation Template}
\label{asec:prompt_template}
We use GPT-4o~\cite{openai2023gpt4} to judge if the model's answer is correct. The specific prompt template with the structured output class is shown below.

\begin{tcolorbox}[
    colback=black!5, colframe=black!50, 
    fontupper=\ttfamily\small, 
    title=Verification Prompts
]
Please verify the following answer:

Question: {question}
Ground Truth Answers: {ground truth}
Predicted Answer: {answer}

Your task is to determine whether the predicted answer correctly matches the ground truth. Focus on overall correctness and provide a detailed explanation in the following format:

class VerificationResult:

    explanation: str   \# Justification 
    
    confidence: float  \# Confidence score in the range [0,1]
    
    correct\_answer: bool  \# True if the prediction is correct, otherwise False
\end{tcolorbox}

\end{document}